\documentclass[final]{cvpr}

\usepackage{times}
\usepackage{epsfig}
\usepackage{graphicx}
\usepackage{amsmath}
\usepackage{amssymb}

\usepackage{balance}

\usepackage{bm}
\newcommand{\matr}[1]{\mathbf{#1}}
\usepackage[pagebackref=true,breaklinks=true,colorlinks,bookmarks=false]{hyperref}

\def\cvprPaperID{4} 
\def\confYear{CVPR 2021}

\begin{document}

\title{
Skeleton Aware Multi-modal Sign Language Recognition
}

\author{Songyao Jiang\textsuperscript{\textsection}, Bin Sun\textsuperscript{\textsection}, Lichen Wang, Yue Bai, Kunpeng Li and Yun Fu \\
Northeastern University, Boston MA, USA\\

}

\maketitle

\begingroup\renewcommand\thefootnote{\textsection}
\footnotetext{Equal contribution}
\endgroup
\begingroup
\renewcommand\thefootnote{}
\footnotetext{This work was supported by the U.S. Army Research Office Award W911NF-17-1-0367.}
\endgroup

\begin{abstract}
Sign language is commonly used by deaf or speech impaired people to communicate but requires significant effort to master. Sign Language Recognition (SLR) aims to bridge the gap between sign language users and others by recognizing signs from given videos. It is an essential yet challenging task since sign language is performed with the fast and complex movement of hand gestures, body posture, and even facial expressions. 
Recently, skeleton-based action recognition attracts increasing attention due to the independence between the subject and background variation. 
However, skeleton-based SLR is still under exploration due to the lack of annotations on hand keypoints. Some efforts have been made to use hand detectors with pose estimators to extract hand key points and learn to recognize sign language via Neural Networks, but none of them outperforms RGB-based methods. 
To this end, we propose a novel Skeleton Aware Multi-modal SLR framework (SAM-SLR) to take advantage of multi-modal information towards a higher recognition rate. Specifically, we propose a Sign Language Graph Convolution Network (SL-GCN) to model the embedded dynamics and a novel Separable Spatial-Temporal Convolution Network (SSTCN) to exploit skeleton features. 
RGB and depth modalities are also incorporated and assembled into our framework to provide global information that is complementary to the skeleton-based methods SL-GCN and SSTCN. 
As a result, SAM-SLR achieves the highest performance in both RGB (98.42\%) and RGB-D (98.53\%) tracks in 2021 Looking at People Large Scale Signer Independent Isolated SLR Challenge. Our code is available at 
\url{https://github.com/jackyjsy/CVPR21Chal-SLR} 
\end{abstract}

\section{Introduction}
Sign language~\cite{SL_intro} is a visual language performed with the dynamic movement of hand gestures, body posture, and facial expressions. It is an effective and helpful approach for deaf and speech-impaired people to communicate with others. Understanding and utilizing sign language requires a considerable time of learning and training which is not practical and feasible for the public. Moreover, sign language is affected by the language~\cite{SL_book1,SL_book2,yang2010chinese} (e.g., English and Chinese) and culture~\cite{different_culture} which further limits its popularization potential. As machine learning and computer vision achieved great progress in the past decade, it is important to explore sign language recognition (SLR) which automatically interprets sign language and helps deaf-mute people communicate smoothly with others in their daily lives.

\begin{figure}[t]
\centering
   \includegraphics[width=0.99\linewidth]{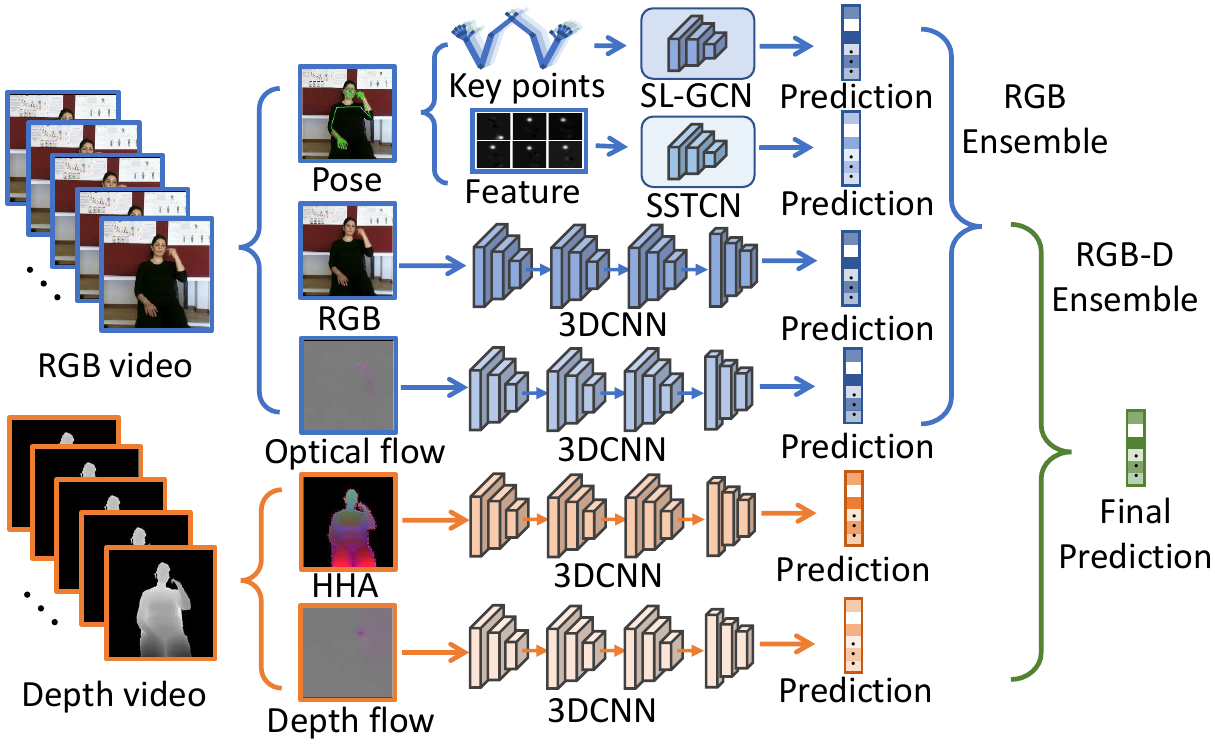}
   \caption{Concept of our Skeleton Aware Multi-modal Sign Language Recognition Framework (SAM-SLR). All local and global motion information is extracted and utilized for final prediction.}
\label{fig:pipline}
\end{figure}


Compared with conventional action recognition, SLR is a more challenging problem. 
First, sign language requires both global body motion and delicate arm/hand gestures to distinctly and accurately express its meaning. Facial expression can be utilized to express emotions as well. 
Similar gestures can even impose various meanings depending on the number of repetitions. 
Second, different signers may perform sign language differently (e.g., speed, localism, left-hander, right-hander, body shape), making SLR more challenging. Collecting more samples from as many signers as possible is desired yet expensive.

Traditional SLR methods mainly deploy handcrafted features such as HOG~\cite{hog_feature} and SIFT~\cite{sift_feature}) associated with conventional classifiers like kNN and SVM~\cite{yang2010chinese,dardas2011real,memics2013kinect}. 
As deep learning achieves significant progress, general video and time-series representation learning methods (e.g., RNN, LSTM) and effective action recognition frameworks (e.g., 3D CNNs) are first explored for SLR tasks in~\cite{sincan2019isolated,li2020word,pigou2018beyond,tur2019isolated}. To more effectively capture the local motion information, attention modules are combined with other modules for higher accuracy~\cite{huang2018video,huang2018attention}. 
Besides, \cite{lim2019isolated,li2018deep,shi2018american} use semantic detection/segmentation models to explicitly guide the recognition network in a two-stage pipeline. 

Recently, skeleton-based methods have become popular in action recognition tasks \cite{yan2018spatial,skeleton_tsrnn,shi2020skeleton,chengdecoupling,song2020stronger} and draw increasing attention due to their strong adaptability to the dynamic circumstances and complicated background. As the skeleton-based methods provide complementary information to the RGB modality, their ensemble results further improve the overall performance. 
However, some deficiencies hinder their extension to the SLR task. Those skeleton-based action recognition methods rely on ground-truth skeleton annotations provided by motion capture systems, restricting themselves to fewer available datasets captured in controlled environments. Besides, most motion capture systems only consider the main body coordinates that do not provide ground truth annotations for hands. Those skeleton data contains insufficient information to recognize sign language, which contains dynamic motions of hand gestures interacted with other body parts. \cite{xiao2020skeleton} attempts to obtain body skeleton and hand poses using separate models and proposes using an RNN-based model to recognize the sign language. But their obtained hand poses are unreliable, and the RNN base model cannot properly model the human skeleton dynamics. 

Inspired by the recent development on whole-body pose estimation \cite{jin2020whole}, in this paper, we propose a novel skeleton-based SLR method using whole-body keypoints and features provided by pretrained whole-body pose estimators. We design a new spatio-temporal skeleton graph for SLR and propose a Sign Language Graph Convolution Network (SL-GCN) to model the dynamics embedded. To fully exploit the information in whole-body keypoints, we propose a novel Separable Spatial-Temporal Convolution Network (SSTCN) for the whole-body skeleton features. Studies on action recognition have revealed that multi-modal data complement each other and provide extra information in recognition. 
To further improve the recognition rate, we propose a Skeleton Aware Multi-modal SLR framework (SAM-SLR) to ensemble the proposed skeleton-based method with other modalities in both RGB and RGB-D scenarios. Our main contributions can be summarized as follows:

\begin{itemize}
\item We construct a novel skeleton graph designed for SLR using whole-body keypoints and graph reduction. Our method utilizes pretrained whole-body pose estimator and requires no extra annotation effort. 

\item We propose SL-GCN to extract information from the whole-body skeleton graph. To our best knowledge, this is the first successful attempt to tackle the SLR task using whole-body skeleton graphs.  

\item We propose a novel SSTCN to further exploit whole-body skeleton features, which can significantly improve the accuracy on whole-body keypoints comparing with the traditional 3D convolution.

\item We propose a SAM-SLR framework for RGB and RGB-D based SLR, which learns from six modalities and achieves the state-of-the-art performance in AUTSL dataset. 
Our proposed method ranked 1st in both RGB and RGB-D tracks in the CVPR-21 Challenge on Isolated SLR~\cite{Sincan:CVPRW:2021}.
\end{itemize}

\section{Related work}
\noindent\textbf{Sign Language Recognition (SLR)} achieves significant progress and obtained high recognition accuracy in recently years due to the development on practical deep learning architectures and the surge of computational power~\cite{sincan2019isolated,li2020word,pigou2018beyond,tur2019isolated,huang2018video,huang2018attention,lim2019isolated,li2018deep,shi2018american}. One remaining challenge of SLR is to capture global body motion information and local arm/hand/facial expression simultaneously. \cite{neverova2015moddrop} proposes a multi-scale and multi-modal framework which captures spatial information at particular spatial scales. An autoencoder framework with connectionist-based recognition module is proposed in \cite{pu2019iterative} for sequence modelling. \cite{koller2018deep} introduces an end-to-end embedding of a convolutional module into a Hidden-Markov-Models, while interpreting the prediction results in a Bayesian framework.  \cite{huang2018attention} proposes a 3D-convolutional neural network associated with attention module which learns the spatio-temporal features from raw video. \cite{pigou2018beyond} incorporates bidirectional recurrence and temporal convolutions together which demonstrates the effectiveness of temporal information in gesture related tasks. \cite{sincan2019isolated} utilizes CNN, Feature Pooling Module, and LSTM Networks associated with adaptive weights to obtain distinctive representations. \cite{huang2018video} designs a Hierarchical Attention Network with Latent Space to eliminate the preprocessing of temporal segmentation. However, these methods mainly consider pure visual feature which is not effective enough to explicitly capture the body movement and hand gesture. \cite{li2020word} designs a pose-based temporal graph convolution networks that model spatial and temporal dependencies in human pose trajectories. \cite{cui2019deep} adopts deep CNNs with stacked temporal fusion layers as the feature extraction module, and bidirectional RNNs as the sequence learning module. \cite{guo2018hierarchical} proposes a hierarchical-LSTM (HLSTM) autoencoder model with visual content and word embedding for translation. It tackles different granularities by conveying spatio-temporal transitions among frames, clips and viseme units. These methods are still not effective enough to capture the complete motion information. 

\noindent\textbf{Skeleton Based Action Recognition} mainly focuses on exploring distinctive patterns from human joint position and motion. Skeleton data can be utilized individually to perform efficient action recognition \cite{cai2021jolo, du2015hierarchical, huang2017deep, li2019actional, li2018independently, liu2016spatio}. On the other hand, it can also be collaborated with other modalities to achieve multi-modal learning aiming for higher recognition performances~\cite{baradel2017human, carreira2017quo, choutas2018potion, hu2018deep, zolfaghari2017chained}. 
RNNs are once popular for modeling skeleton data~\cite{du2015hierarchical, liu2016spatio, li2018independently, si2018skeleton}. 
Recently, \cite{yan2018spatial} is the first attempt to designed a graph-based approach, called ST-GCN, to model the dynamic patterns in skeleton data via a Graph Convolutional Network (GCN). Such approach draws much attention and a few improvements have been developed as well \cite{li2019actional, shi2019skeleton, shi2019two, si2019attention, shi2020skeleton, chengdecoupling, song2020stronger}. Specifically,
\cite{li2019actional} propose a AS-GCN to dig the latent joint connections to boost the recognition performance. A two-stream approach is presented in \cite{shi2019two} and further extended to four streams in \cite{shi2020skeleton}. \cite{chengdecoupling} develops a decoupling GCN to increase the model capacity with no extra computational cost. ResGCN is proposed in \cite{song2020stronger} which adopts a bottleneck structure from ResNet \cite{he2016deep} to reduce parameters while increasing model capacity. 
However, skeleton based SLR is still under-explored. An attempt to extend ST-GCN to SLR directly \cite{de2019spatial} has been unsuccessful that only achieves around 60\% recognition rate on 20 sign classes, which is significantly lower than handcrafted features. 
\label{sec:related_work}
\noindent\textbf{Multi-modal Approach} aims to explore action data captured from different resources/devices to improve the final performance. It is based on the assumption that different modalities contain unique motion information which could potentially complement each other and eventually obtain comprehensive and distinctive action representations. 
View-invariant representation learning framework is proposed in \cite{zheng2016cross,wang2013dense} to obtain robust representation for down-stream tasks. \cite{hoff_CVPR16} deploys shared weights network on multi-modal scenario to obtain modality hallucination for image classification task. DA-Net~\cite{Wang_2018_ECCV} proposes a view-specific and a view-independent modules to capture the features and effectively merges the prediction scores together. 
A feature factorization framework is proposed in \cite{shahroudy2018deep} which explores the view shared-specific information for RGB-D action recognition. A cascaded residual autoencoder is designed in~\cite{tran_CVPR17} to handle incomplete view classification setting. A super vector is proposed in \cite{multiview_action2} to fuse the multi-view representations together. \cite{GMVAR_Lichen} proposes a cross-view generative strategy to explore the latent view distribution connections and a late fusion strategy to effectively learn the prediction correlations. 
Encouraged by the success of those multi-modal methods, we aim to explore more visual, depth, gesture, and hand modalities jointly to capture information from all aspects and fuse them together via a universal framework to achieve higher performance.

\begin{figure*}[!ht]
  \centering
  \includegraphics[width=0.98\textwidth]{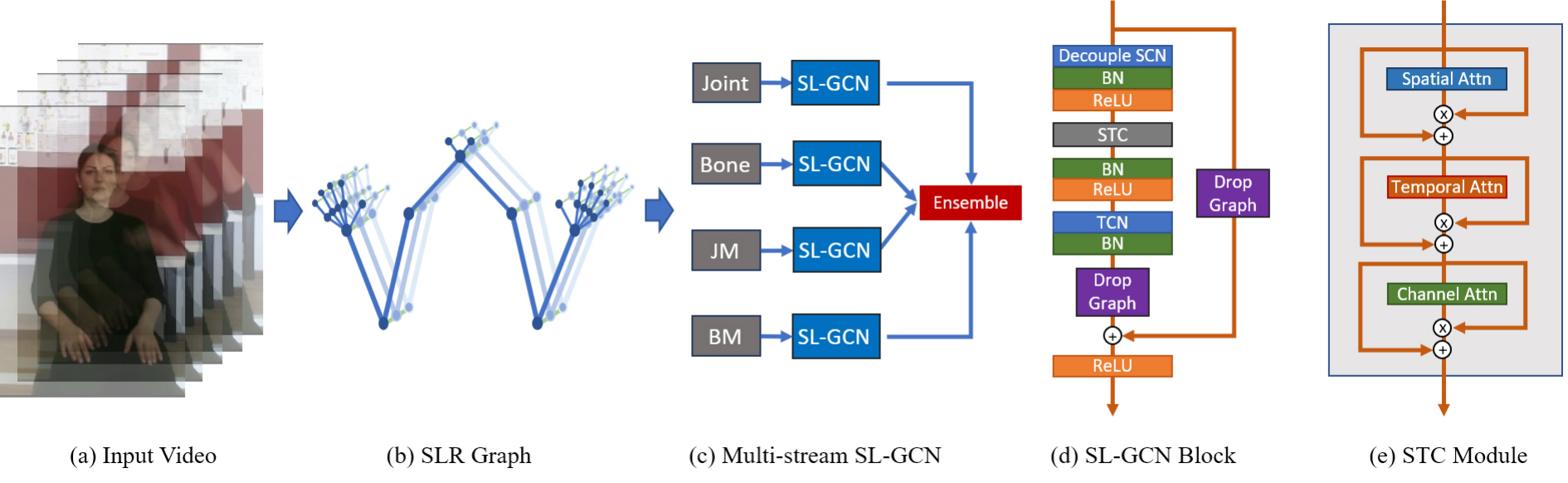}
  \caption{Illustration of the SL-GCN pipeline: (a) Input sign language videos; (b) SLR graph constructed from whole-body keypoints after graph reduction; (c) Workflow of the multi-stream SL-GCN (Joint, Bone, JM=Joint Motion, BM=Bone Motion); (d) SL-GCN block architecture; (e) STC attention module used in the SL-GCN block. }
\label{fig:gcn-pipeline}
\end{figure*}

\section{Our approach}
This section will first introduce SL-GCN and SSTCN models based on skeleton keypoints and features, respectively. Then we will present a baseline 3D CNNs model for other modalities. Last, we will introduce our SAM-SLR framework and explain the multi-modal ensemble process.

\subsection{SL-GCN}

We construct a spatio-temporal graph to model the dynamics of human body skeleton for SLR, and propose a SL-GCN model with attention mechanism to extract motion dynamics from the graph. We also adopt a multi-stream approach to further improve the performance.

\subsubsection{Graph Construction and Reduction}
\label{sec:graph_construction}
Hand gestures play an important role in performing sign language. For action recognition, researchers tend to use the ground-truth skeleton annotations provided by motion capture system such as Kinect v2 \cite{kinect_v1v2}. Unfortunately, such system does not provide annotations for the hands. We use a pretrained whole-body pose estimation network to provide 133 keypoints estimated from the detected person in videos. A spatio-temporal graph can then be constructed by connecting the adjacent keypoints in the spatial dimension according to the natural connections of human body, and connecting all keypoints to themselves in the temporal dimension. In this graph, the node set $V=\{v_{i,t}|i=1,...,N,t=1,...,T\}$ includes all facial landmarks, body skeleton, hands, and feet keypoints. Then an adjacent matrix $\matr{A}$ in spatial dimension can be constructed as
\begin{equation}
    \matr{A}_{i,j}=
        \begin{cases}
      1 & \text{if $d(v_i, v_j)=1$}\\
      0 & \text{else}
    \end{cases} 
    \label{eqn:graph_construct}
\end{equation}
where $d(v_i, v_j)$ calculate the minimum distance (the minimum number of nodes in the shortest path) between skeleton node $v_i$ and $v_j$. 

However, different from the graph used in action recognition which contains a small number of nodes, the large number of nodes and edges in the whole-body skeleton graph introduces a lot of noise to the model. Besides, if two nodes are far away with many nodes in between, it is difficult to learn the interactions between those nodes. Simply using such whole-body skeleton graph containing all the 133 nodes gives a low accuracy in our experiment. Therefore, based on observations on the data and visualizations of GCN activations, we conduct a graph reduction on the whole-body skeleton graph and trim down the 133 nodes to 27 nodes. The remaining graph contains 10 nodes for each hand and 7 nodes for the upper body, which is illustrated in Figure \ref{fig:gcn-pipeline}(b). Our experiments demonstrate that such graph contain the essential information we need for SLR. Graph reduction results in faster model convergence and significantly higher recognition rate.


\subsubsection{Graph Convolution}
To capture the pattern embedded in the whole-body skeleton graph, we adopt the spatio-temporal GCN in \cite{yan2018spatial} with spatial partitioning strategy to model the dynamic skeletons. The implementation of spatial GCN can be expressed as
\begin{equation}
    \matr{x}_\text{out}=\matr{D}^{-\frac{1}{2}}(\matr{A}+\matr{I})\matr{D}^{-\frac{1}{2}}\matr{x}_\text{in}\matr{W},
    \label{eqn:graph_conv}
\end{equation}
where adjacent matrix $\matr{A}$ represents intra-body connections and an identity matrix $\matr{I}$ represents self-connections, $\matr{D}$ presents the diagonal degree of $(\matr{A}+\matr{I})$, and $\matr{W}$ is a trainable weight matrix of the convolution. In practice, such GCN is implemented as performing standard 2D convolution and then multiplying the results by $\matr{D}^{-\frac{1}{2}}(\matr{A}+\matr{I})\matr{D}^{-\frac{1}{2}}$. The temporal GCN can be also implemented as a standard 2D convolution with kernel size $k_t \times 1$ that it performs on the temporal dimension with a reception field $k_t$. 
We adopt a extended variation of the spatial graph convolution called decoupling graph convolution proposed in \cite{chengdecoupling} to further boost the capacity of GCN. In decoupling graph convolution, the channels of graph features split into $G$ groups and channels in each group share an independent trainable adjacent matrix $\matr{A}$. The convolution results of the decoupling groups are concatenated together as the output feature.

\subsubsection{SL-GCN Block}
Our proposed SL-GCN Block is constructed with decoupled spatial convolutional network, self-attention and graph dropping module inspired by \cite{shi2020skeleton,chengdecoupling}. As illustrated in Figure \ref{fig:gcn-pipeline}(d), a basic GCN block of our proposed SL-GCN network consists of a decoupled spatial convolutional layer (Decouple SCN), a STC (spatial, temporal and channel-wise) attention module, a temporal convolutional layer (TCN) and a DropGraph module. The Decouple SCN boosts the GCN modeling capacity without extra cost. The DropGraph module avoid overfitting. The STC attention mechanism consists of a spatial attention module, a temporal attention module and a channel attention module connected in a cascaded configuration, as illustrated in Figure \ref{fig:gcn-pipeline}(e). Our proposed spatio-temporal GCN consists of 10 such GCN blocks. At the end, a global average pooling is applied on both spatial and temporal dimensions before classification using a fully-connected layer. 

\subsubsection{Multi-stream Approach}
\label{sec:multi-stream}
Inspired by \cite{shi2020skeleton}, 1st-order representation (joints coordinate), 2nd-order representation (bone vector), and their motion vectors are worth to be investigated for SLR. As shown in Figure \ref{fig:gcn-pipeline}(c), our multi-stream SL-GCN uses joint, bone, joint motion, and bone motion. Bone data are generated by representing joint data in a vector form pointing from source joints to their target joints following the natural connections of human body. The nose node is used as the root joint so that its bone vector is assigned to be zeros. Let the source and target joint be represented as $v^{J}_{i,t}=(x_{i,t},y_{i,t},s_{i,t})$ and $v^{J}_{j,t}=(x_{j,t},y_{j,t},s_{j,t})$ where $(x,y,s)$ represents x-y coordinates and confidence score, the bone vectors of the other nodes can be calculated by subtracting their source joint coordinates from their current joint coordinates as $v^B_{j,t}=(x_{j,t}-x_{i,t},y_{j,t}-y_{i,t},s_{j,t})$, for all $(i,j)\in H$ where $H$ is the set of naturally connected human body. Motion data are generated by calculating the difference between adjacent frames in both joint and bone streams. Joint motion can be calculated as $v^{JM}_{i,t}=(x_{i,t+1}-x_{i,t},y_{i,t+1}-y_{i,t},s_{i,t})$ and bone motion can be calculated as $v^{BM}_{i,t}=v^B_{i,t+1}-v^B_{i,t}$. We train each stream separately and combine their predicted results by adding them together with weights using the same ensemble strategy described in Section \ref{sec:ensemble}. We tried to adopt an early fused multi-stream method proposed in ResGCN \cite{song2020stronger} which captures multi-stream features via multiple input blocks capture and concatenates them together afterwards. However, our implementation does not provide better performance, so we stick to the late ensemble method and leave it to be explored in future work.

\begin{figure}
  \centering
  \includegraphics[width=0.48\textwidth]{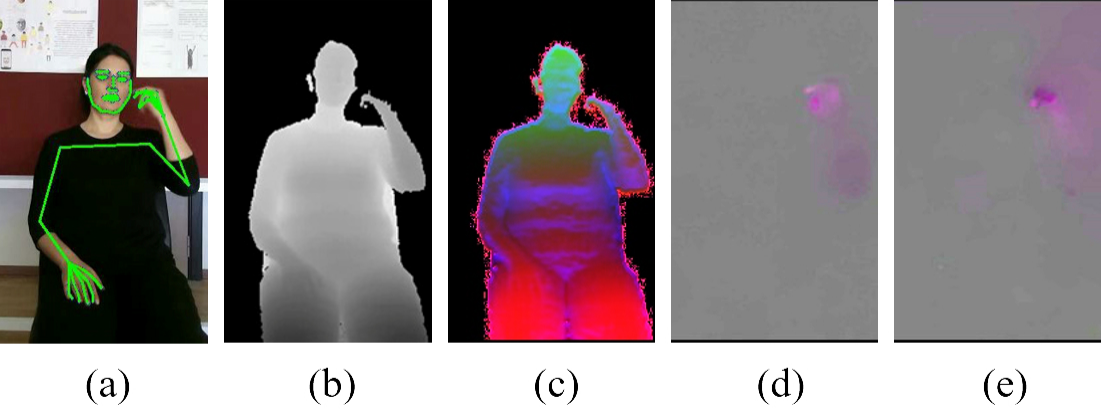}
  \caption{Visualization of modalities: (a) RGB with whole-body keypoints overlay; (b) Depth; (c) Masked HHA; (d) Optical flow; (e) Depth flow. (better viewed in color)}
\label{fig:modalities}
\end{figure}

\subsection{SSTCN}
\label{sec:SSTCN}

\begin{figure}[t]
  \centering
  \includegraphics[width=0.47\textwidth]{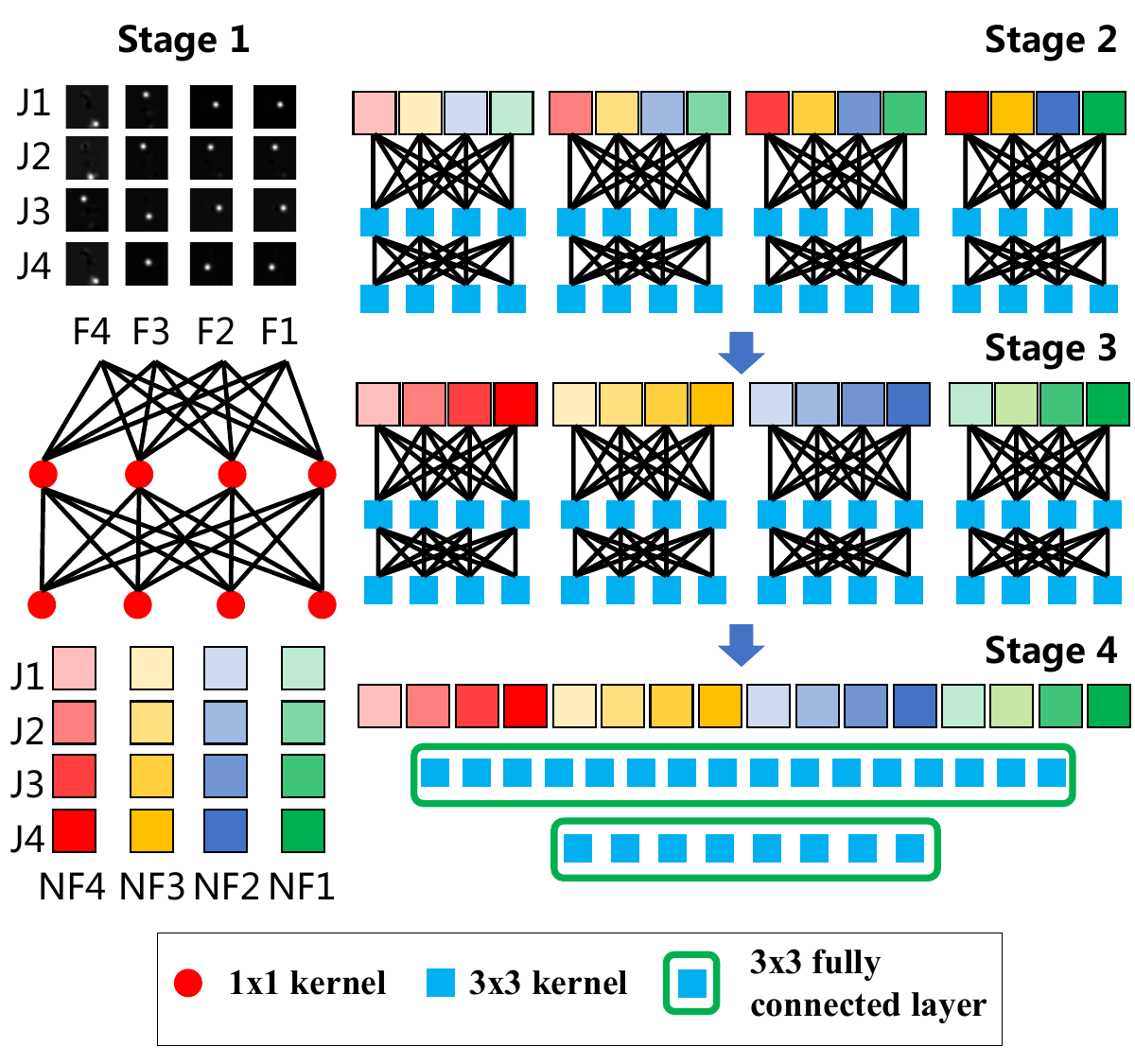}
  \caption{Illustration of the proposed SSTCN for skeleton features. Abbrevs: J=Joints; F=Frames; NF=New Features.}
  \label{fig:tpose}
\end{figure}

Besides using key point coordinates generated from the whole-body pose network, we also propose a SSTCN model to recognize the sign language from whole-body features. We extract features of 33 keypoints from 60 frames of each video as the input to our model, which contain 1 landmark on the nose, 4 landmarks on mouth, 2 landmarks on shoulders, 2 landmarks on elbows, 2 landmarks on wrists, and 22 landmarks on hands. All the features are down-sampled to $24 \times 24$ using max pooling. Instead of using 3D convolution, we process the input features with a 2D convolution layer separably, which reduces the parameters and makes it easier to converge. The pipeline is shown in Figure \ref{fig:tpose}. There are four stages in total. In the first stage, we reshape the features from $60\times33\times24\times24$ to $60\times792\times24$, and feed them to $1 \times 1$ convolution layers, which means we only process temporal information in this stage. 
Then we shuffle the features and divide them into $60$ groups, and utilize grouped $3 \times 3$ convolution to extract temporal and spatial information among the same key point features from different frames. In this stage, temporal information and part of spatial information are processed. In the third stage, the features are shuffled again and divided into $33$ groups. We still use grouped $3 \times 3$ convolution, but only spatial information in each frame is extracted. Finally, a couple of $3 \times 3$ fully connected layers are implemented to generate prediction features. In the first 3 stages, all the output is added by a residual. Moreover, a dropout layer is deployed in each module to avoid over-fitting \cite{dropout}. An ablation study on the effectiveness of SSTCN is shown in Section \ref{sec:ablation study SSTCN}. 
To further improve the performance, we utilize the Swish \cite{swish} activation function, which can be written as:
\begin{equation}
    f(x)=x \cdot \text{Sigmoid}(x).
    \label{eqn:swish}
\end{equation}

Since using one-hot labels with cross-entropy loss results in overfitting in some cases \cite{ren2015faster}, we adopt the label smoothing technique to alleviate such effect. Mathematically, label smoothing can be represented as 
\begin{equation}
    q'(k|x)=(1-\epsilon)\delta_{k,y}+\epsilon u(k),
    \label{eqn:label_smooth}
\end{equation}
where $q'(k|x)$ is a new form of predicted distribution, $\epsilon$ is a hyper-parameter between $0$ and $1$, $u()$ is a uniform distribution and $k$ is the number of classes. The cross-entropy loss can then be replaced as

\begin{equation}
    H(q',p)=-\sum_{k=1}^K \log p(k)q'(k) = (1-\epsilon) H(q,p) + \epsilon H(u,p),
    \label{eqn:cross_entropy}
\end{equation}
where such representation can be regarded as a combination of penalties to the difference between the predicted distribution with the real distribution and the prior distribution (uniform distribution). 


\subsection{3D CNNs}
As mentioned in Section~\ref{sec:related_work}, studies on action recognition have revealed that multi-modal ensembles can further boost each modality's performance, hence we construct a simple but effective baseline using 3D CNNs for the other modalities of RGB frames, optical flows, depth HHA and depth flow. 
In our study, we find out that ResNet2+1D \cite{tran2018closer}, which decouples spatial and temporal convolution in 3D CNNs and does them one after another, provides the best result among popular 3D CNN architectures. We find that increasing the architecture depth does not improve the performance and makes the network easier to overfit. So in our experiment, we choose ResNet2+1D-18 with weights pretrained on Kinectics dataset \cite{carreira2018short} as the backbone network. To further improve the recognition rate, for RGB frames, we pretrain the model on the Chinese Sign Language (CSL) dataset \cite{zhang2016chinese}. We find that pretraining on CSL can improve the model convergence and increase the final accuracy by around $1\%$. Same as SSTCN, we replace the ReLU activations with Swish activations (Equation~\ref{eqn:swish}) and use the label smoothing technique with corresponding cross-entropy loss in Equation~\ref{eqn:label_smooth} and \ref{eqn:cross_entropy} to avoid overfitting.

\subsection{Multi-modal Ensemble}
\label{sec:ensemble}
We use a simple ensemble method to ensemble all four modalities above. Specifically, we save the output of the last fully-connected layers of each modality before softmax layer. Those outputs have the size $n_c$ where $n_c$ is the number of classes. We assign weights to the every modality according to their accuracy on validation set and sum them up with weights as our final predicted score
\begin{equation}
    q_\text{RGB} = \alpha_1 q_\text{skel} + \alpha_2 q_\text{RGB}  + \alpha_3 q_\text{flow} + \alpha_4 q_\text{feat},
\end{equation}
\begin{align}
\begin{aligned}
    q_\text{RGB-D} = & \alpha_1 q_\text{skel} +  \alpha_2 q_\text{RGB}  + \alpha_3 q_\text{flow} + \alpha_4 q_\text{feat}\\
    &  + \alpha_5 q_\text{HHA} + \alpha_6 q_\text{depthflow},
\end{aligned}
\end{align}
where $q$ represents the result of each modality, $\alpha_{1,2,3,4,5,6}$ are hyper-parameters to be tuned based on ensemble accuracy on validation set. We find the indices of maximum scores as our final predicted classes using an $\text{argmax}()$ operator. In our experiments, we use $\alpha = [1,0.9,0.4,0.4]$ for RGB track and $\alpha = [1.0,0.9,0.4,0.4,0.4,0.1]$ for RGB-D track. We have tried other ensemble methods such as early fusion or training fully-connected layers to ensemble the final prediction. Despite that, we find that the simplest method we presented above gives us the best accuracy. 
\section{Experiments}
In this section, we present evaluation of our proposed SAM-SLR framework on the AUTSL dataset. We start from a brief introduction about the AUTSL dataset and how we extract the data of all modalities. Then we evaluate our single-modal models and using the validation data compared with the baseline methods. Besides, we demonstrate that the effectiveness of proposed approaches via ablation studies on the model components. After that, we fuse the results from different modalities in both RGB and RGB-D scenarios to show that those modalities complement each other and improve the overall recognition rate. Last, we show our evaluated results on the test set, which ranked the 1st place in the SLR challenge~\cite{Sincan:CVPRW:2021}. 

\subsection{AUTSL Dataset}
AUTSL~\cite{SLR_dataset} is collected for general SLR tasks in Turkish sign language. Kinect V2 sensor~\cite{kinect_v1v2,kinect2_acc} is utilized in the collection procedure. Specifically, 43 signers with 20 backgrounds are assigned to perform 226 different sign actions. In general, it contains 38,336 video clips which is split to training, validation, and testing subsets. 
The statistical summary of the balanced dataset, which is used in the challenge, is listed in Table ~\ref{dataset_table}.


\subsection{Baseline Methods}
Along with the AUTSL benchmark \cite{SLR_dataset}, several deep learning based models are proposed. We treat the best model benchmarked in \cite{SLR_dataset} as well as the SLR challenge leader board as the baseline model here (Baseline RGB and Baseline RGB-D in Table~\ref{tab:validation_results_all}). Specifically, the model is mainly constructed using CNN + LSTM structure, where 2D-CNN model are used to extract feature for each video frame and bidirectional LSTMs (BLSTM) are adopted on top of the these 2D CNN features to lean their temporal relations. A feature pooling model (FPM) \cite{sincan2019isolated} is plugged in after the 2D CNN model to obtain multi-scale representation of the features. A spacial-temporal attention model \cite{raffel2015feed} is then built on top of BLSTM features to better focus on important spacial-temporal information for SLR.

\begin{table}
\begin{center}
\begin{tabular}{l c r}
\hline
Subsets & Signers & Samples \\
\hline
Training & 31 & 28,142\\
Validation & 5 & 4,418\\
Testing & 7 & 3,742\\
Total & 43 & 36,302\\
\hline
\end{tabular}
\end{center}\caption{A statistical summary of the balanced AUTSL dataset.}\label{dataset_table}
\end{table}


\subsection{Multi-modal Data Preparation}

\noindent\textbf{Whole-body Pose Keypoints and Features}. 
We use a pretrained HRNet \cite{sun2019deep} whole-body pose estimator provided by MMPose \cite{mmpose2020} to estimate 133-point whole-body keypoints from the RGB videos and construct the 27-node skeleton graph in Section \ref{sec:graph_construction}. As mentioned in Section \ref{sec:multi-stream}, we process the graph into four streams (joint, bone, joint motion and bone motion). 
Randomly sampling, mirroring, rotating, scaling, jittering and shifting are applied as data augmentations. We use a sample length of 150 in our experiment. If a video has lesser frames than 150, we repeat that video until we get 150 frames. Coordinates of keypoints are normalized to [-1,1]. For skeleton features, as mentioned in \ref{sec:SSTCN}, we choose $33$ joint features for each frame. $60$ frames will be uniformly sampled from each video.

\noindent\textbf{RGB Frames and Optical Flow}. 
All frames of RGB videos are extracted and saved as images for faster parallel loading and processing. 
We follow the same process in~\cite{wang2016temporal} to obtain optical flow features using TVL1 algorithm~\cite{zach2007duality} implemented with OpenCV and CUDA. The output flow maps of x and y directions are concatenated in channel dimension. RGB frames and optical flow frames are cropped and resized to $256 \times 256$ based on the keypoints obtained from whole-body pose estimation. Such cropping and resizing operations are performed on the other image-like modalities as well. During training, we randomly sample 32 consecutive frames for each video. When testing, we uniformly sample 5 such clips from input videos and average on their predicted score.

\noindent\textbf{Depth HHA and Depth Flow}. 
We extract HHA features from depth videos as another modality. HHA features encode depth information into a RGB-like 3-channel output, where HHA stand for horizontal disparity, height above the ground and angle normal. 
Using HHA instead of using gray-scale depth videos directly enables better understanding of the scene and improves the recognition rate. We observe that the provided depth videos come with a mask. So when generating HHA features, we mask out those regions and fill them with zeros. An example of our extracted HHA with mask can be found in Figure \ref{fig:modalities}(c). Black regions are masked out pixels. We treat HHA feature the same way as the RGB frames in data augmentation. 
Besides, we follow the exactly the same procedure used for RGB to extract optical flow from the depth modality (named depth flow). The depth flow is cleaner and captures different information compared with the RGB flow, as shown in Figure \ref{fig:modalities}(e). 
  
\begin{table}
\begin{center}
\begin{tabular}{l | c | c}
\hline
Streams & Top-1 & Top-5 \\
\hline
\hline
Joint       & 95.02 & 99.21\\
Bone        & 94.70 & 99.14\\
Joint Motion& 93.01 & 98.85\\
Bone Motion & 92.49 & 98.78\\
\hline
Multi-stream &\textbf{95.45} & \textbf{99.25}\\
\hline
\end{tabular}
\end{center}\caption{Performance of multi-stream SL-GCN on validation set.}
\label{tab:validation_results_keypoints}
\end{table}

\begin{table}
\begin{center}
\begin{tabular}{l | c}
\hline
Variations & Top-1 \\
\hline
\hline
SL-GCN (Joint) & \textbf{95.02} \\
\hline
w/o Graph Reduction & 63.69\\
w/o Decouple GCN & 94.66\\
w/o Drop Graph & 94.81\\
w/o Keypoints Augmentation & 90.16\\
w/o STC Attention & 93.53\\
\hline
\end{tabular}
\end{center}\caption{Ablation studies on SL-GCN using joint stream.}
\label{tab:ablation_gcn}
\end{table}

\subsection{Performance of SL-GCN}
The results of our proposed SL-SLR are reported in Table \ref{tab:validation_results_keypoints} in terms of Top-1 and Top-5 recognition rate. The joint stream provides the best performance among all four streams, and their ensemble further improve the overall recognition rate, which demonstrates the effectiveness of our proposed multi-stream SL-GCN model using whole-body skeleton graph. Our SL-GCN performs the best among the other single-modality models as shown in Table \ref{tab:validation_results_all}. Another major advantage of the graph based method is that it is much faster to run compared with 3D CNNs using RGB frames, since the data is less complex and requires lower computational operations.

Ablation studies on the proposed SL-GCN model is presented in Table \ref{tab:ablation_gcn}. Our graph reduction technique is the most significant contributor to the performance. Without the graph reduction, the GCN model can hardly learn from the complex dynamics in the skeleton graph with too many nodes and edges. The data augmentation techniques (i.e., random sampling, mirror, rotate, shift, jitter) are also crucial in learning the dynamics embedded, since the GCN model is easy to overfit on the data. The decoupling GCN module, the DropGraph module and the STC attention mechanism all contribute to our final recognition rate as well. 
 
\begin{table}
\begin{center}
\begin{tabular}{l | c | c}
\hline
Methods & Feature size & Top-1 \\
\hline
\hline
ResNet3D       & $12\times12$ & 92.82\\
ResNet2+1D     & $12\times12$ & 93.03\\
\textbf{SSTCN} & $12\times12$ & \textbf{93.60}\\
\hline
\textbf{SSTCN} &$24\times24$ & \textbf{94.32}\\
\hline
\end{tabular}
\end{center}\caption{Comparing our SSTCN with ResNet3D and ResNet2+1D on $12\times12$ feature size shows the effectiveness of our SSTCN. Using larger feature size will further improve the performance.}
\label{table:comparison_SSTCN}
\end{table}


\subsection{SSTCN Results}
\label{sec:ablation study SSTCN}
In this subsection, the details of training will be presented. Besides, we will show the comparison results with ResNet3D~\cite{hara2018can,kataoka2020would} and ResNet2+1D~\cite{tran2018closer} to show the effectiveness of our model. As shown in Figure \ref{fig:tpose}, our model has 4 stages in total. Each stage has two layers. Since the last two layers are fully connected layers that may significantly impact performance due to their redundant parameters, we utilize three ResNet3D and ResNet2+1D modules while implementing ResNet3D ResNet2+1D, respectively. The training loss and the position of dropout layers are mentioned in Section~\ref{sec:SSTCN}. The learning rate is $1e-3$ in the beginning with weight decay $1e-4$. At epoch 50, the learning rate is set as $1e-4$, and the weight decay is set to 0. At epoch 100, the learning rate is set as $1e-5$. We trained 200 epochs in total. The hyper-parameters remain the same while training ResNet3D and ResNet2+1D as baselines. We also compare the results of different feature sizes. The comparison results are shown in Table \ref{table:comparison_SSTCN}. From the table, we can find out that our SSTCN has the highest accuracy comparing with ResNet3D and ResNet2+1D on the same scale of features. With a larger feature size, our SSTCN can achieve even better performance. 

\begin{table}
\begin{center}
\begin{tabular}{l | c}
\hline
3D CNN Variations & Top-1 \\
\hline
\hline
Ours (RGB Frame) & \textbf{94.77} \\
\hline
w/o Label Smoothing & 93.75\\
w/o Swish Activation & 92.88\\
w/o Pretraining on CSL & 93.41\\
w/ ResNet3D-18 Backbone & 93.10\\
\hline
\end{tabular}
\end{center}\caption{Ablation studies on 3D CNN using RGB frames.}
\label{tab:ablation_rgb}

\end{table}

\begin{table}[t]
\begin{center}
\begin{tabular}{l | c | c}
\hline
Modality & Top-1 & Top-5 \\
\hline
\hline
Baseline RGB & 42.58 & -\\
Baseline RGB-D & 63.22 & -\\
\hline
Keypoints       & \textbf{95.45} & \textbf{99.25}\\
Features        & 94.32 & 98.84\\
RGB Frames      & 94.77 & 99.48\\
RGB Flow        & 91.65 & 98.76\\
Depth HHA       & 95.13 & 99.25\\
Depth Flow      & 92.69 & 98.87\\
\hline
\end{tabular}\end{center}\caption{Results of single modalities on AUTSL validation set.}
\label{tab:validation_results_all}
\end{table}

\subsection{Other Modalities and Ensembles}
The results of our baseline 3D CNNs for the the other modalities are reported in Table \ref{tab:validation_results_all}. Keypoints method represents our proposed multi-stream SL-GCN, which performs the best among the other single-modality methods. If we consider the feature based method using SSTCN as the same modality as SL-GCN (both skeleton based), their ensemble result achieves even higher recognition rate, see Table \ref{tab:multi-modal}. 
We observe that the depth flow provides slightly better accuracy compared with RGB flow due to the lesser noise introduced. An ablation study on the 3D CNN architecture is also provided in Table \ref{tab:ablation_rgb} using the RGB frames. From the ablation study, we find that label smoothing and swish activation both improve the recognition rate by 1\% and 2\%, respectively. Pretraining on CSL dataset \cite{zhang2016chinese} improves the final accuracy by 1.4\%. 

The ensemble results in both RGB and RGB-D scenarios using different combinations of modalities are summarized in Table~\ref{tab:multi-modal} as two groups. The skeleton based method combined from SL-GCN and SSTCN performs better than RGB + Flow and Depth ensemble models, which shows the effectiveness of our proposed skeleton based approach. The ensemble results of RGB All and RGB-D All demonstrate that the whole-body skeleton based approaches are able to collaborate with the other modalities and further improve the final recognition rate. 

\subsection{Evaluated on Challenge Test Set}
When training our models on the training set, we adopt an early stopping technique based on the validation accuracy to obtain our best models. Then we test our best models on the test set and use the hyperparameters tuned on validation set to obtain our ensemble prediction. 
In the final test phase of the challenge, we are allowed to finetune our model using the validation set. To further improve our performance, we finetune our best models on the union of training and validation set. Since we cannot validate the training models this time, we stop training when the training loss in our finetuning experiment is reduced to the same level as our best models in the training phase. Our predictions with and without finetuning are evaluated on the challenge server and reported in Table \ref{tab:test_results}. Our proposed SAM-SLR approach surpasses the baseline methods significantly and ranked 1st in both RGB and RGB-D tracks of the SLR challenge. 

\begin{table}
\small{
\begin{center}
\begin{tabular}{l | c | c | c | c | c | c | c | c}
\hline
Ensemble & K & F & R & O & H & D & Top-1 & Top-5\\
\hline
\hline
Skeleton  & \checkmark & \checkmark &&&&& 96.11 & 99.43 \\
RGB+Flow &&& \checkmark & \checkmark &&& 95.77 & 99.52\\
RGB All & \checkmark & \checkmark & \checkmark & \checkmark &&& 96.96 & 99.68 \\
\hline
Depth &&&&& \checkmark & \checkmark & 95.76 & 99.41\\
RGB+D && & \checkmark & \checkmark & \checkmark & \checkmark & 96.27 & 99.66 \\
RGBD All & \checkmark & \checkmark & \checkmark & \checkmark & \checkmark & \checkmark & \textbf{97.10} & \textbf{99.73} \\
\hline
\end{tabular}\end{center}}\caption{Multi-modal ensemble results evaluated on AUTSL validation set. Abbrevs: K=Keypoints; F=Features; R=RGB; O=Optical Flow; H=HHA; D=Depth Flow.}
\label{tab:multi-modal}
\end{table}

\begin{table}[t]
\begin{center}
\begin{tabular}{l | c | l | c }
\hline
  & Finetune & Track & Top-1 \\
\hline
\hline
Baseline & - & RGB  & 49.23\\
Baseline & - & RGB-D & 62.03\\
\hline
Ensemble & No & RGB & 97.51\\
Ensemble & No & RGB-D & 97.68\\
\hline
Ensemble & w/ Val & RGB     & \textbf{98.42}\\
Ensemble & w/ Val & RGB-D   & \textbf{98.53}\\
\hline
\end{tabular}
\end{center}\caption{Performance our ensemble results (with and without finetuning) evaluated on AUTSL test set.}
\label{tab:test_results}
\end{table}

\section{Conclusion}


In this paper, we propose a novel Skeleton Aware Multi-modal SLR framework (SAM-SLR) to take advantage of multi-modal information towards effective SLR. Specifically, we construct a skeleton graph for SLR using pretrained whole-body pose estimators and propose SL-GCN to model the embedded spatial and temporal dynamics. Our approach requires no extra effort on skeleton annotation. In addition to modeling keypoints dynamics, we propose SSTCN to exploit information in skeleton features. Furthermore, we implement effective baselines for the other RGB and depth modalities and assemble all modalities together in the proposed SAM-SLR framework, which achieves the state-of-the-art performance and won the challenge on SLR in both RGB and RGB-D tracks. We hope our work could encourage and facilitate future research on SLR.

    

{\small
\balance
\bibliographystyle{ieee_fullname}
\bibliography{egbib}

\begin{thebibliography}{10}\itemsep=-1pt

\bibitem{kinect2_acc}
Clemens Amon, Ferdinand Fuhrmann, and Franz Graf.
\newblock Evaluation of the spatial resolution accuracy of the face tracking
  system for {K}inect for windows {V}1 and {V}2.
\newblock In {\em Proceedings of AAAI Conference on Artificial}, pages 16--17,
  2014.

\bibitem{baradel2017human}
Fabien Baradel, Christian Wolf, and Julien Mille.
\newblock Human action recognition: Pose-based attention draws focus to hands.
\newblock In {\em Proceedings of the IEEE International Conference on Computer
  Vision Workshops}, pages 604--613, 2017.

\bibitem{cai2021jolo}
Jinmiao Cai, Nianjuan Jiang, Xiaoguang Han, Kui Jia, and Jiangbo Lu.
\newblock Jolo-gcn: Mining joint-centered light-weight information for
  skeleton-based action recognition.
\newblock In {\em Proceedings of the IEEE/CVF Winter Conference on Applications
  of Computer Vision}, pages 2735--2744, 2021.

\bibitem{multiview_action2}
Zhuowei Cai, Limin Wang, Xiaojiang Peng, and Yu Qiao.
\newblock Multi-view super vector for action recognition.
\newblock In {\em Proceedings of IEEE Computer Vision and Pattern Recognition},
  pages 596--603, 2014.

\bibitem{carreira2018short}
Joao Carreira, Eric Noland, Andras Banki-Horvath, Chloe Hillier, and Andrew
  Zisserman.
\newblock A short note about kinetics-600.
\newblock {\em arXiv preprint arXiv:1808.01340}, 2018.

\bibitem{carreira2017quo}
Joao Carreira and Andrew Zisserman.
\newblock Quo vadis, action recognition? a new model and the kinetics dataset.
\newblock In {\em proceedings of the IEEE Conference on Computer Vision and
  Pattern Recognition}, pages 6299--6308, 2017.

\bibitem{chengdecoupling}
Ke Cheng, Yifan Zhang, Congqi Cao, Lei Shi, Jian Cheng, and Hanqing Lu.
\newblock Decoupling {GCN} with {DropGraph} module for skeleton-based action
  recognition.
\newblock In {\em Proceedings of European Conference on Computer Vision}, 2020.

\bibitem{choutas2018potion}
Vasileios Choutas, Philippe Weinzaepfel, J{\'e}r{\^o}me Revaud, and Cordelia
  Schmid.
\newblock Potion: Pose motion representation for action recognition.
\newblock In {\em Proceedings of the IEEE conference on computer vision and
  pattern recognition}, pages 7024--7033, 2018.

\bibitem{mmpose2020}
MMPose Contributors.
\newblock {OpenMMLab} pose estimation toolbox and benchmark.
\newblock \url{https://github.com/open-mmlab/mmpose}, 2020.

\bibitem{cui2019deep}
Runpeng Cui, Hu Liu, and Changshui Zhang.
\newblock A deep neural framework for continuous sign language recognition by
  iterative training.
\newblock {\em IEEE Transactions on Multimedia}, 21(7):1880--1891, 2019.

\bibitem{dardas2011real}
Nasser~H Dardas and Nicolas~D Georganas.
\newblock Real-time hand gesture detection and recognition using
  bag-of-features and support vector machine techniques.
\newblock {\em IEEE Transactions on Instrumentation and measurement},
  60(11):3592--3607, 2011.

\bibitem{de2019spatial}
Cleison~Correia de Amorim, David Mac{\^e}do, and Cleber Zanchettin.
\newblock Spatial-temporal graph convolutional networks for sign language
  recognition.
\newblock In {\em International Conference on Artificial Neural Networks},
  pages 646--657. Springer, 2019.

\bibitem{du2015hierarchical}
Yong Du, Wei Wang, and Liang Wang.
\newblock Hierarchical recurrent neural network for skeleton based action
  recognition.
\newblock In {\em Proceedings of the IEEE conference on computer vision and
  pattern recognition}, pages 1110--1118, 2015.

\bibitem{SL_intro}
Karen Emmorey.
\newblock {\em Language, cognition, and the brain: Insights from sign language
  research}.
\newblock Psychology Press, 2001.

\bibitem{guo2018hierarchical}
Dan Guo, Wengang Zhou, Houqiang Li, and Meng Wang.
\newblock Hierarchical {LSTM} for sign language translation.
\newblock In {\em Proceedings of AAAI Conference on Artificial Intelligence},
  volume~32, 2018.

\bibitem{hara2018can}
Kensho Hara, Hirokatsu Kataoka, and Yutaka Satoh.
\newblock Can spatiotemporal {3D CNN}s retrace the history of {2D CNN}s and
  {ImageNet}?
\newblock In {\em Proceedings of IEEE Computer Vision and Pattern Recognition},
  pages 6546--6555, 2018.

\bibitem{he2016deep}
Kaiming He, Xiangyu Zhang, Shaoqing Ren, and Jian Sun.
\newblock Deep residual learning for image recognition.
\newblock In {\em Proceedings of the IEEE conference on computer vision and
  pattern recognition}, pages 770--778, 2016.

\bibitem{hoff_CVPR16}
Judy Hoffman, Saurabh Gupta, and Trevor Darrell.
\newblock Learning with side information through modality hallucination.
\newblock In {\em Proceedings of IEEE Computer Vision and Pattern Recognition},
  pages 826--834, 2016.

\bibitem{hu2018deep}
Jian-Fang Hu, Wei-Shi Zheng, Jiahui Pan, Jianhuang Lai, and Jianguo Zhang.
\newblock Deep bilinear learning for rgb-d action recognition.
\newblock In {\em Proceedings of the European Conference on Computer Vision
  (ECCV)}, pages 335--351, 2018.

\bibitem{huang2018attention}
Jie Huang, Wengang Zhou, Houqiang Li, and Weiping Li.
\newblock Attention-based {3D-CNN}s for large-vocabulary sign language
  recognition.
\newblock {\em IEEE Transactions on Circuits and Systems for Video Technology},
  29(9):2822--2832, 2018.

\bibitem{huang2018video}
Jie Huang, Wengang Zhou, Qilin Zhang, Houqiang Li, and Weiping Li.
\newblock Video-based sign language recognition without temporal segmentation.
\newblock In {\em Proceedings of AAAI Conference on Artificial Intelligence},
  volume~32, 2018.

\bibitem{huang2017deep}
Zhiwu Huang, Chengde Wan, Thomas Probst, and Luc Van~Gool.
\newblock Deep learning on lie groups for skeleton-based action recognition.
\newblock In {\em Proceedings of the IEEE conference on computer vision and
  pattern recognition}, pages 6099--6108, 2017.

\bibitem{jin2020whole}
Sheng Jin, Lumin Xu, Jin Xu, Can Wang, Wentao Liu, Chen Qian, Wanli Ouyang, and
  Ping Luo.
\newblock Whole-body human pose estimation in the wild.
\newblock In {\em Proceedings of European Conference on Computer Vision}, 2020.

\bibitem{SL_book2}
Trevor Johnston and Adam Schembri.
\newblock {\em Australian Sign Language (Auslan): An introduction to sign
  language linguistics}.
\newblock Cambridge University Press, 2007.

\bibitem{kataoka2020would}
Hirokatsu Kataoka, Tenga Wakamiya, Kensho Hara, and Yutaka Satoh.
\newblock Would mega-scale datasets further enhance spatiotemporal {3D CNN}s?
\newblock {\em arXiv preprint arXiv:2004.04968}, 2020.

\bibitem{koller2018deep}
Oscar Koller, Sepehr Zargaran, Hermann Ney, and Richard Bowden.
\newblock Deep sign: {E}nabling robust statistical continuous sign language
  recognition via hybrid {CNN-HMM}s.
\newblock {\em International Journal of Computer Vision}, 126(12):1311--1325,
  2018.

\bibitem{li2020word}
Dongxu Li, Cristian Rodriguez, Xin Yu, and Hongdong Li.
\newblock Word-level deep sign language recognition from video: {A} new
  large-scale dataset and methods comparison.
\newblock In {\em Proceedings of IEEE Winter Conference on Applications of
  Computer Vision}, pages 1459--1469, 2020.

\bibitem{li2019actional}
Maosen Li, Siheng Chen, Xu Chen, Ya Zhang, Yanfeng Wang, and Qi Tian.
\newblock Actional-structural graph convolutional networks for skeleton-based
  action recognition.
\newblock In {\em Proceedings of the IEEE/CVF Conference on Computer Vision and
  Pattern Recognition}, pages 3595--3603, 2019.

\bibitem{li2018independently}
Shuai Li, Wanqing Li, Chris Cook, Ce Zhu, and Yanbo Gao.
\newblock Independently recurrent neural network (indrnn): Building a longer
  and deeper rnn.
\newblock In {\em Proceedings of the IEEE conference on computer vision and
  pattern recognition}, pages 5457--5466, 2018.

\bibitem{li2018deep}
Yuan Li, Xinggang Wang, Wenyu Liu, and Bin Feng.
\newblock Deep attention network for joint hand gesture localization and
  recognition using static {RGB-D} images.
\newblock {\em Information Sciences}, 441:66--78, 2018.

\bibitem{lim2019isolated}
Kian~Ming Lim, Alan Wee~Chiat Tan, Chin~Poo Lee, and Shing~Chiang Tan.
\newblock Isolated sign language recognition using convolutional neural network
  hand modelling and hand energy image.
\newblock {\em Multimedia Tools and Applications}, 78(14):19917--19944, 2019.

\bibitem{liu2016spatio}
Jun Liu, Amir Shahroudy, Dong Xu, and Gang Wang.
\newblock Spatio-temporal lstm with trust gates for 3d human action
  recognition.
\newblock In {\em European conference on computer vision}, pages 816--833.
  Springer, 2016.

\bibitem{sift_feature}
David~G Lowe.
\newblock Object recognition from local scale-invariant features.
\newblock In {\em Proceedings of IEEE International Conference on Computer
  Vision}, volume~2, pages 1150--1157, 1999.

\bibitem{memics2013kinect}
Abbas Memi{\c{s}} and Song{\"u}l Albayrak.
\newblock A {K}inect based sign language recognition system using
  spatio-temporal features.
\newblock In {\em Proceedings of International Conference on Machine Vision},
  volume 9067, page 90670X. International Society for Optics and Photonics,
  2013.

\bibitem{different_culture}
Anna Mindess.
\newblock {\em Reading between the signs: Intercultural communication for sign
  language interpreters}.
\newblock Nicholas Brealey, 2014.

\bibitem{neverova2015moddrop}
Natalia Neverova, Christian Wolf, Graham Taylor, and Florian Nebout.
\newblock {ModDrop}: {A}daptive multi-modal gesture recognition.
\newblock {\em IEEE Transactions on Pattern Analysis and Machine Intelligence},
  38(8):1692--1706, 2015.

\bibitem{kinect_v1v2}
Diana Pagliari and Livio Pinto.
\newblock Calibration of kinect for {X}box {O}ne and comparison between the two
  generations of microsoft sensors.
\newblock 15:27569--27589, 10 2015.

\bibitem{pigou2018beyond}
Lionel Pigou, A{\"a}ron Van Den~Oord, Sander Dieleman, Mieke Van~Herreweghe,
  and Joni Dambre.
\newblock Beyond temporal pooling: {R}ecurrence and temporal convolutions for
  gesture recognition in video.
\newblock {\em International Journal of Computer Vision}, 126(2):430--439,
  2018.

\bibitem{pu2019iterative}
Junfu Pu, Wengang Zhou, and Houqiang Li.
\newblock Iterative alignment network for continuous sign language recognition.
\newblock In {\em Proceedings of IEEE Conference on Computer Vision and Pattern
  Recognition}, pages 4165--4174, 2019.

\bibitem{raffel2015feed}
Colin Raffel and Daniel~PW Ellis.
\newblock Feed-forward networks with attention can solve some long-term memory
  problems.
\newblock {\em arXiv:1512.08756}, 2015.

\bibitem{swish}
Prajit Ramachandran, Barret Zoph, and Quoc~V Le.
\newblock Searching for activation functions.
\newblock {\em arXiv preprint arXiv:1710.05941}, 2017.

\bibitem{ren2015faster}
Shaoqing Ren, Kaiming He, Ross Girshick, and Jian Sun.
\newblock Faster {R-CNN}: {T}owards real-time object detection with region
  proposal networks.
\newblock {\em arXiv preprint arXiv:1506.01497}, 2015.

\bibitem{shahroudy2018deep}
Amir Shahroudy, Tian-Tsong Ng, Yihong Gong, and Gang Wang.
\newblock Deep multimodal feature analysis for action recognition in {RGB+D}
  videos.
\newblock {\em IEEE Transactions on Pattern Analysis and Machine Intelligence},
  40(5):1045--1058, 2018.

\bibitem{shi2018american}
Bowen Shi, Aurora~Martinez Del~Rio, Jonathan Keane, Jonathan Michaux, Diane
  Brentari, Greg Shakhnarovich, and Karen Livescu.
\newblock American sign language fingerspelling recognition in the wild.
\newblock In {\em IEEE Spoken Language Technology Workshop}, pages 145--152,
  2018.

\bibitem{shi2019skeleton}
Lei Shi, Yifan Zhang, Jian Cheng, and Hanqing Lu.
\newblock Skeleton-based action recognition with directed graph neural
  networks.
\newblock In {\em Proceedings of the IEEE/CVF Conference on Computer Vision and
  Pattern Recognition}, pages 7912--7921, 2019.

\bibitem{shi2019two}
Lei Shi, Yifan Zhang, Jian Cheng, and Hanqing Lu.
\newblock Two-stream adaptive graph convolutional networks for skeleton-based
  action recognition.
\newblock In {\em Proceedings of the IEEE/CVF Conference on Computer Vision and
  Pattern Recognition}, pages 12026--12035, 2019.

\bibitem{shi2020skeleton}
Lei Shi, Yifan Zhang, Jian Cheng, and Hanqing Lu.
\newblock Skeleton-based action recognition with multi-stream adaptive graph
  convolutional networks.
\newblock {\em IEEE Transactions on Image Processing}, 29:9532--9545, 2020.

\bibitem{si2019attention}
Chenyang Si, Wentao Chen, Wei Wang, Liang Wang, and Tieniu Tan.
\newblock An attention enhanced graph convolutional lstm network for
  skeleton-based action recognition.
\newblock In {\em Proceedings of the IEEE/CVF Conference on Computer Vision and
  Pattern Recognition}, pages 1227--1236, 2019.

\bibitem{si2018skeleton}
Chenyang Si, Ya Jing, Wei Wang, Liang Wang, and Tieniu Tan.
\newblock Skeleton-based action recognition with spatial reasoning and temporal
  stack learning.
\newblock In {\em Proceedings of the European Conference on Computer Vision
  (ECCV)}, pages 103--118, 2018.

\bibitem{Sincan:CVPRW:2021}
Ozge~Mercanoglu Sincan, Julio C.~S. Jacques~Junior, Sergio Escalera, and
  Hacer~Yalim Keles.
\newblock Chalearn {LAP} large scale signer independent isolated sign language
  recognition challenge: Design, results and future research.
\newblock In {\em Proceedings of the IEEE/CVF Conference on Computer Vision and
  Pattern Recognition (CVPR) Workshops}, 2021.

\bibitem{SLR_dataset}
Ozge~Mercanoglu Sincan and Hacer~Yalim Keles.
\newblock {AUTSL}: {A} large scale multi-modal turkish sign language dataset
  and baseline methods.
\newblock {\em IEEE Access}, 8:181340--181355, 2020.

\bibitem{sincan2019isolated}
Ozge~Mercanoglu Sincan, Anil~Osman Tur, and Hacer~Yalim Keles.
\newblock Isolated sign language recognition with multi-scale features using
  {LSTM}.
\newblock In {\em 2019 27th Signal Processing and Communications Applications
  Conference (SIU)}, pages 1--4. IEEE, 2019.

\bibitem{song2020stronger}
Yi-Fan Song, Zhang Zhang, Caifeng Shan, and Liang Wang.
\newblock Stronger, faster and more explainable: {A} graph convolutional
  baseline for skeleton-based action recognition.
\newblock In {\em Proceedings of ACM International Conference on Multimedia},
  pages 1625--1633, 2020.

\bibitem{dropout}
Nitish Srivastava, Geoffrey Hinton, Alex Krizhevsky, Ilya Sutskever, and Ruslan
  Salakhutdinov.
\newblock Dropout: a simple way to prevent neural networks from overfitting.
\newblock {\em Journal of Machine Learning Research}, 15(1):1929--1958, 2014.

\bibitem{sun2019deep}
Ke Sun, Bin Xiao, Dong Liu, and Jingdong Wang.
\newblock Deep high-resolution representation learning for human pose
  estimation.
\newblock In {\em Proceedings of IEEE Computer Vision and Pattern Recognition},
  pages 5693--5703, 2019.

\bibitem{tran2018closer}
Du Tran, Heng Wang, Lorenzo Torresani, Jamie Ray, Yann LeCun, and Manohar
  Paluri.
\newblock A closer look at spatiotemporal convolutions for action recognition.
\newblock In {\em Proceedings of IEEE Computer Vision and Pattern Recognition},
  pages 6450--6459, 2018.

\bibitem{tran_CVPR17}
Luan Tran, Xiaoming Liu, Jiayu Zhou, and Rong Jin.
\newblock Missing modalities imputation via cascaded residual autoencoder.
\newblock In {\em Proceedings of IEEE Computer Vision and Pattern Recognition},
  pages 1405--1414, 2017.

\bibitem{tur2019isolated}
Anil~Osman Tur and Hacer~Yalim Keles.
\newblock Isolated sign recognition with a siamese neural network of {RGB} and
  depth streams.
\newblock In {\em IEEE International Conference on Smart Technologies}, pages
  1--6, 2019.

\bibitem{SL_book1}
Clayton Valli and Ceil Lucas.
\newblock {\em Linguistics of American sign language: An introduction}.
\newblock Gallaudet University Press, 2000.

\bibitem{Wang_2018_ECCV}
Dongang Wang, Wanli Ouyang, Wen Li, and Dong Xu.
\newblock Dividing and aggregating network for multi-view action recognition.
\newblock In {\em Proceedings of European Conference on Computer Vision},
  September 2018.

\bibitem{wang2013dense}
Heng Wang, Alexander Kl{\"a}ser, Cordelia Schmid, and Cheng-Lin Liu.
\newblock Dense trajectories and motion boundary descriptors for action
  recognition.
\newblock {\em International Journal of Computer Vision}, 103(1):60--79, 2013.

\bibitem{skeleton_tsrnn}
Hongsong Wang and Liang Wang.
\newblock Modeling temporal dynamics and spatial configurations of actions
  using two-stream recurrent neural networks.
\newblock In {\em Proceedings of IEEE Computer Vision and Pattern Recognition},
  2017.

\bibitem{GMVAR_Lichen}
Lichen Wang, Zhengming Ding, Zhiqiang Tao, Yunyu Liu, and Yun Fu.
\newblock Generative multi-view human action recognition.
\newblock In {\em Proceedings of IEEE International Conference on Computer
  Vision}, pages 6212--6221, 2019.

\bibitem{wang2016temporal}
Limin Wang, Yuanjun Xiong, Zhe Wang, Yu Qiao, Dahua Lin, Xiaoou Tang, and Luc
  Van~Gool.
\newblock Temporal segment networks: Towards good practices for deep action
  recognition.
\newblock In {\em Proceedings of European Conference on Computer Vision}, pages
  20--36. Springer, 2016.

\bibitem{xiao2020skeleton}
Qinkun Xiao, Minying Qin, and Yuting Yin.
\newblock Skeleton-based chinese sign language recognition and generation for
  bidirectional communication between deaf and hearing people.
\newblock {\em Neural Networks}, 125:41--55, 2020.

\bibitem{yan2018spatial}
Sijie Yan, Yuanjun Xiong, and Dahua Lin.
\newblock Spatial temporal graph convolutional networks for skeleton-based
  action recognition.
\newblock In {\em Proceedings of AAAI conference on artificial intelligence},
  volume~32, 2018.

\bibitem{yang2010chinese}
Quan Yang.
\newblock Chinese sign language recognition based on video sequence appearance
  modeling.
\newblock In {\em Proceedings of IEEE Conference on Industrial Electronics and
  Applications}, pages 1537--1542, 2010.

\bibitem{zach2007duality}
Christopher Zach, Thomas Pock, and Horst Bischof.
\newblock A duality based approach for realtime {TV-L}1 optical flow.
\newblock In {\em Proceedings of Joint Pattern Recognition Symposium}, pages
  214--223. Springer, 2007.

\bibitem{zhang2016chinese}
Jihai Zhang, Wengang Zhou, Chao Xie, Junfu Pu, and Houqiang Li.
\newblock Chinese sign language recognition with adaptive {HMM}.
\newblock In {\em Proceedings of IEEE International Conference on Multimedia
  and Expo}, pages 1--6, 2016.

\bibitem{zheng2016cross}
Jingjing Zheng, Zhuolin Jiang, and Rama Chellappa.
\newblock Cross-view action recognition via transferable dictionary learning.
\newblock {\em IEEE Transactions on Image Processing}, 25(6):2542--2556, 2016.

\bibitem{hog_feature}
Qiang Zhu, Mei-Chen Yeh, Kwang-Ting Cheng, and Shai Avidan.
\newblock Fast human detection using a cascade of histograms of oriented
  gradients.
\newblock In {\em Proceedings of IEEE Computer Vision and Pattern Recognition},
  volume~2, pages 1491--1498, 2006.

\bibitem{zolfaghari2017chained}
Mohammadreza Zolfaghari, Gabriel~L Oliveira, Nima Sedaghat, and Thomas Brox.
\newblock Chained multi-stream networks exploiting pose, motion, and appearance
  for action classification and detection.
\newblock In {\em Proceedings of the IEEE International Conference on Computer
  Vision}, pages 2904--2913, 2017.

\end{thebibliography}
}

\end{document}